\documentclass{article}
\usepackage{spconf,amsmath,graphicx}
\usepackage{blindtext}

\usepackage{color}
\definecolor{cardinal}{cmyk}{0,1,0.63,0.29}

\newcommand{\eg}{\textit{e.g.,\ }}
\newcommand{\etc}{\textit{etc.\@}}

\newcommand{\etal}{\textit{et al.\@}}

\newcommand{\Pg}[1]{\noindent\emph{\textbf{\color{cardinal}#1}}}%paragraph

\title{Unsupervised Latent Behavior Manifold Learning from Acoustic Features: audio2behavior}
\name{Haoqi Li$^1$, Brian Baucom$^2$, Panayiotis Georgiou$^1$\thanks{Work supported by NSF and DoD.}}
\address{$^1$University of Southern California, Los Angeles, CA, USA \\
  $^2$The University of Utah, Department of Psychology, UT, USA \\}

\begin{document}
\ninept
\maketitle
\begin{abstract}
Behavioral annotation using signal processing and machine learning is highly dependent on training data and manual annotations of behavioral labels. Previous studies have shown that speech information encodes significant behavioral information and be used in a variety of automated behavior recognition tasks. However, extracting behavior information from speech is still a difficult task due to the sparseness of training data coupled with the complex, high-dimensionality of speech, and the complex and multiple information streams it encodes. In this work we exploit the slow varying properties of human behavior. We hypothesize that nearby segments of speech share the same behavioral context and hence share a similar underlying representation in a latent space. Specifically, we propose a Deep Neural Network (DNN) model to connect behavioral context and derive the behavioral  manifold  in an unsupervised manner. We evaluate the proposed manifold in the couples therapy domain and also provide examples from publicly available data (e.g. stand-up comedy). We further investigate training within the couples' therapy domain and from movie data. The results are extremely encouraging and promise improved behavioral quantification in an unsupervised manner and warrants further investigation in a range of applications.
 
\end{abstract}
\section{Introduction}
\label{sec:intro}
Analysis and classification of human behaviors is one of the core tasks of observational study. For example, in couples therapy, psychologists observe and identify domain-specific behaviors(\eg blame and acceptance) during couple interactions, and provide specific treatments based on their analysis. 

Behavior estimation process is a complicated task. Different from emotions, human behaviors such as acceptance, are often manifested over long time scales. Longer context needs to be considered when human annotators attempt to quantify behavior. Because of that, human raters need to combine information at different timescales to estimate behaviors correctly. It is difficult to simulate the complex non-linear nature of the annotation process using one specific algorithm. Moreover, data with rich behavioral information from psychotherapy domains are often severely limited in quantity due to privacy constraints and cost of annotation.

Integrating machine learning and signal processing methods, Behavior Signal Processing (BSP)\cite{intro_BSP, Georgiou:2011:BSP} employs acoustic\cite{Black20131, xia2015dynamic}, lexical\cite{Georgiou2011, tseng2016_couples-behavio}, and visual\cite{7106538, 6607640} information to model and analyze multi-modal human behaviors. For example, in couples therapy domain, using acoustic features, Black \etal \cite{Black20131} built an automatic human behavioral coding system for couples’ interaction. To deal with data sparsity, a sparsely connected and disjointly trained deep neural networks (SD-DNN) framework was introduced in \cite{li2016_sparsely-connec}, that limits the number of trained parameters at any time. 

Despite these efforts in the BSP domain, it is still challenging to extract effective behavior representations from high-dimensionality acoustic features. Over the last few years, Deep Neural Networks have demonstrated promise in their capability to learn high level representation from raw data. For instance, by training DNN with audio features input, and corresponding labels(\eg emotion recognition in \cite{han2014speech, le2013emotion}, keyword spotting in \cite{chen2015query}) as target, the output of DNN can be regarded as representation of raw input data. However, this supervised framework fails in our specific domain, since a huge amount of training data with annotated labels is essential. Data sparseness limits the use of AI methods for emotions, stress, and behavior estimation. Thus, in this work we propose an  unsupervised way of exploiting data for the BSP domain. We further investigate whether out of domain data can be employed for in-domain behavioral quantification.

Recently, context information has been used for a range of applications. For instance in developing the word2vec model Milikov \etal \cite{mikolov2013distributed,mikolov2013efficient} have proposed an embedding that ties 1-hot word representations of nearby words via an intermediate, hidden, vector representation. Similar to auto-encoders or bottleneck representations\cite{hinton2006reducing, baldi2012autoencoders}, the hidden layer attempts to connect the information at the input and output layers, but in this case the information resides at a longer scale than either of the two representations -- namely context.

Our proposed framework employs a similar idea to the word2vec. Since humans employ a large temporal window to observe the context and evaluate behaviors, we can hence assume that behavior remains relatively constant within a sufficiently long window. This matches also annotation guidelines in the field of psychology where the minimum observation windows are usually set at 30 seconds. It also matches empirical understanding of behavior. For example, one person (often the case in couples therapy interactions as well as everyday life) can be sad  during a conversion for a long window of time despite different intonations and speech patterns throughout that temporal window.

%As an empirical rule, the success of machine learning algorithms generally depends on data representation\cite{Bengio:2012wg}. Based on the geometric notion of manifold, an effective behavior manifold should preserve the ``behavioral direction'', while reduce the ``non-behavioral direction''. 

In our paper, we propose an unsupervised behavior manifold learning using Deep Neural Network via unlabeled acoustic features. We learn the manifold with unlabeled within-domain data and from Out-Of-Domain (OOD) data. We evaluate if the knowledge gained includes behaviorally meaningful information within and OOD training and within and OOD testing.

The rest of paper is organized as follows: Section 2 describes in detail our proposed manifold learning to obtain behavior representation in an unsupervised manner. Section 3 provides a brief description of the database used in our paper, after which we describe audio processing, feature extraction steps and experiment settings in section 4. After that, we discuss our results in section 5. Finally, we give our conclusion and future work in section 6.

\section{Methodology}
\label{sec:method}
The success of machine learning algorithms can be attributed to two main properties: first the DNN can represent any function, and second it can learn that function based on large amounts of data. The underlying representations that the DNN identifies are critical to its success\cite{Bengio:2012wg, bengio2012deep}. In the BSP domain, we often suffer from lack of data while the complexities of the signal require the use of high-dimensionality acoustic features. The goal of this paper is to identify, in an unsupervised manner, a latent manifold  where the signal retains its behavioral characteristics. In this behavioral manifold we expect similar behaviors to appear closer together than they do in the original signal space or in the feature space. Based on the geometric notion of manifolds, the learned representation can be associated with an intrinsic coordinate system on the embedded manifold\cite{Bengio:2012wg}. In our case, an effective behavior manifold should preserve information residing on a ``behavioral axis'', while removing other acoustically encoded information. 

One reasonable assumption is that the behavioral state of a person is slow varying (note that behavior changes much slower than emotional expression despite the close relations between the two). This means that by looking at a very short interval of behavior (say 5s) and a following interval (say next 5s), we will most likely observe the same or a very similar behavioral state. Based on this assumption we will create a model that exploits context and ties the two intervals via the proposed reduced dimensionality embedding vector space.

We acknowledge and expect the following complication with the above assumption: the nearby information frames also encode speaker characteristics as well as acoustic conditions such as environment and channel. We will discuss this further in Section \ref{sec:discussion}.

\subsection{Training framework}
\label{ssec:train}

\begin{figure}[t]
  \centering
  \includegraphics[width=\linewidth]{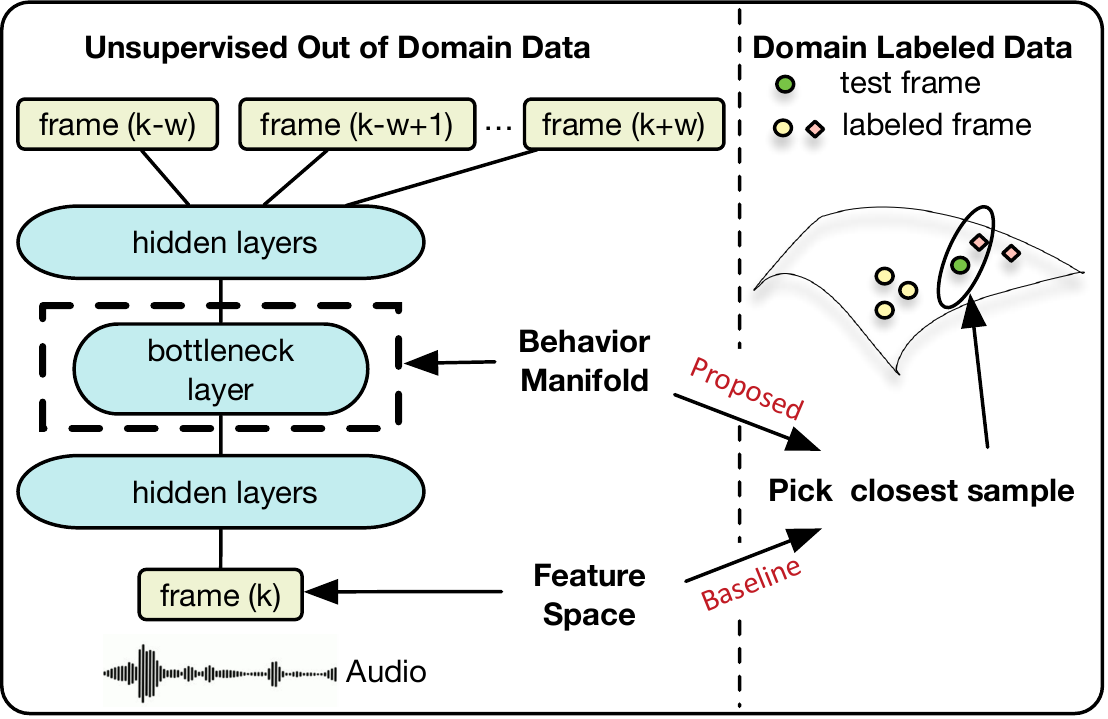}
  \caption{{Behavior representation training framework}}
  \label{fig:OverallProcess}
\end{figure}

Our proposed training framework is similar to an autoencoder, but rather than just training to reconstruct the input our system trains to reconstruct neighboring frames.
As shown in Fig.~\ref{fig:OverallProcess}, for the $kth$ frame of acoustic features, the outputs are frames from $k-w$, $k+w$ excluding the $kth$ frame, where $w$ is the size of the window in which we consider behavioral context to remain relatively constant. By creating such an unsupervised corpus we can train similarly to standard DNN tasks with back propagation, thus learning the underlying behavioral manifold representation.

\subsection{Behavior manifold representation}
\label{ssec:subhead}
After the training, we use the output of the bottleneck layer as the behavior representation. In general, the dimension of the hidden layer is smaller than dimension of the original feature space, so this process can be also regarded as a  feature dimensionality reduction or compression process. 

\subsection{Evaluation}
\label{ssec:subhead}
Since we employ an unsupervised method in training our model, we need to demonstrate that representations indeed include behavior information.  We intend to do this on different evaluation data: (i) From the field of psychology we will employ as a case study Couples Therapy interactions and we will compare underlying representations of similarly rated sessions. For example, after learning manifold on unsupervised data, a test session (with known rating) can be compared in the manifold space with all known samples of negative/positive behaviors, and closest match can be selected. (ii) We also collect a range of samples from political speeches, stand-up comedy \etc and compare their pairwise similarity. Details of the datasets are provided below.

\section{Corpus}
\label{sec:data}
For the unsupervised training process we utilize two corpora in our paper:
\begin{itemize}
\item \emph{Ti:} For in-domain BSP data (Train-in-domain: Ti) we employ the couples therapy database by UCLA/UW Couple Therapy Research Project\cite{christensen2004traditional}, in which 134 couples were involved in video-taped marital issue interactions. In each session, one relationship-related topic (\eg ``Why can't you leave my stuff alone?'') was initiated during the speech session. Although not used for the training, behavioral labels exist for this corpus.
%In this paper, we primarily focus on ``Acceptance'' behavioral code.

\item \emph{To:}  For out of BSP domain training dataset, we collected around 400 hours of audio from a range of movies. Many of the selected movies include large parts of emotional conversions reflecting a range of behaviors.
\end{itemize}
~\\For testing we also employ two datasets:
\begin{itemize}
\item  \emph{Eo:}  For out of domain evaluation  (Eo) data, we collected audio from two different speakers for each of the following scenarios: stand-up comedy of comedians who employ anger as an elicitation mechanism (see Table \ref{tab:EoData}), comedian without angry behavior, political debate, Ted talk, eulogy. Each audio's length is around 10 minutes. \\

\item \emph{Ei:} Within the BSP domain we employ the labels of our couples therapy data. Each participant's behavior was evaluated by trained human annotators for a set of 33 behaviors(\eg ``Acceptance'', ``Blame'' \etc) based on standard Couples Interaction\cite{heavey2002couples} and Social Support Rating Systems\cite{jones1998couples}. Each annotator rated 1-9 for each behavior at session level in terms of the presence of this behavior. In this work we show relationships with 4 of the behaviors by binarizing the top and bottom 20\% of the original ratings.
\end{itemize}
%1-1-comedian_angry-George_Carlin-About_Rape
%1-2-comedian_angry-Richard_pryor-mudbone_little_feets
%2-1-comedian_soft-Jim_Gaffigan_Jesus_Beyond_the_Pale
%2-2-comedian_soft-Steve_Hofstetter-Stand_up_show
%3-1-politician_debate-Final_republican_presidential_debate_of_2015
%3-2-politician_debate-Vice_presidential_debate_2012_Complete_The_Candidates_Debate
%4-1-ted-Kevin_Slavin-How_algorithms_shape_our_world
%4-2-ted-Christopher_Steiner-Algorithms_are_taking_over_the_world
%5-1-eulogy-Eulogy_for_a_Son_from_youtube
%5-2-eulogy-Mr_Li_Hongyi's_eulogy_for_the_late_Mr_Lee_Kuan_Yew

\begin{table}[t]
  \centering
  \caption{Out of Domain test data\label{tab:EoData}}~\\[1ex]
  \hrule~\\[1ex]
  1. George Carlin;
  2. Richard Pryor; 
  3. Jim Gaffigan;
  4. Steve Hofstetter
  5. Final Republican Presidential Debate, 2015
  6. Vice Presidential Debate 2012
  7. TEDtalk: Kevin Slavin;
  8. Christopher Steiner
  9. Eulogy for a Son (youtube)
  10. Mr.\ Li Hongyi's Eulogy for the late Mr.\ Lee Kuan Yew\\[1ex]

  \hrule~\\
\end{table}
%1-1-comedian_angry-George_Carlin-About_Rape.wav
%1-2-comedian_angry-Richard_pryor-mudbone_little_feets.wav
%2-1-comedian_soft-Jim_Gaffigan_Jesus_Beyond_the_Pale.wav
%2-2-comedian_soft-Steve_Hofstetter-Stand_Up.wav
%3-1-politician_debate-FINAL_REPUBLICAN_PRESIDENTIAL_DEBATE_OF_2015.wav
%3-2-politician_debate-VicePresidentialDebate2012Complete.wav
%4-1-tednew-How_algorithmsapeourworldKevinSlavin.wav
%4-2-tednew-AlgorithmsAreTakingOverTheWorld.wav
%5-1-eulogy-EulogyForASon.wav
%5-2-eulogy-Mr_Li_Hongyis_eulogyforthe_late_Mr_Lee_Kuan_Yew.wav

\section{Experiments}
\label{sec:exp}

\subsection{Audio processing}
\label{sec:pre-processing}
For couples therapy data, due to the limitations of the available recordings, some pre-processing is needed to remove sessions with low SNR. Further, since these are dyadic interaction data, we wanted to diarize the interactions. In this work, we employed the same  pre-processing as in\cite{Black20131}. In short, we utilize all interactions with an SNR above 5dB, perform Voice Activity Detection (VAD) to identify spoken regions, and  Speaker Diarization to identify same-speaker regions.

For the movie dataset we did not perform any pre-processing procedure, thus treating all frames the same, including silence, music, and changing speaker regions.

\subsection{Acoustic feature extraction}
\label{sec:feat}
We extract acoustic features characterizing speech prosody (pitch, intensity and their derivatives), spectral envelope characteristics (MFCCs, MFBs, LPCs and their derivatives), voice quality (jitter, shimmer and their derivatives). All of these Low-Level Descriptors (LLD) are extracted every 10ms with a 25ms Hamming window using openSMILE\cite{opensmile}. Within each frame, we compute functionals of all these acoustic features including Min (1st percentile), Max (99th percentile), Range (99th percentile -- 1st percentile), Mean, Median, and Standard Deviation.

Temporal variation of behavior is much slower than basic emotion, thus a longer size of frame window is necessary for its analysis. In our paper, in order to estimate meaningful behavioral metrics while maintaining high resolution, we use a 20s and 5s windows with 1s shift. 

\subsection{Experimental Setup}
We conducted three different types of experiments, using different training or evaluation datasets. In our experiments, the input dimension of our feature is 420 as discussion in section ~\ref{sec:feat}. The dimension of the bottleneck layer is set to be 64. To generate the training data, for each frame, the window range $w$ is set to be 6, we randomly pick up 5 context frames within context window as reconstruction labels for this frame. For instance for an input of audio from 100-120s the context could be any five frames from $(100+i) -(120+i)$ where $i=[-6,-5,\ldots,6]$. 

Three experiments are described as follows:
\begin{itemize}
\item \textbf{Experiment (1):} Unsupervised training on couples therapy corpus (Ti), and evaluate on Couples therapy corpus (Ei). 
\item \textbf{Experiment (2):} Unsupervised training on movie corpus (To), and test on Couples Therapy corpus (Ei).  
\item \textbf{Experiment (3):} Unsupervised training on movie corpus (To), and test on audio sessions listed in table \ref{tab:EoData} representing different behavior styles (To, Eo).
\end{itemize}

\subsection{Evaluation Method for In-Domain Couples Therapy}
\label{sec:IDeva}
As mentioned before, we have only session-level ratings for the couples therapy corpus. For each behavior code and each gender, we selected 70 sessions on one extreme case of this code (\eg high acceptance) and another 70 sessions at the other extreme (\eg low acceptance). We binarize the behavior to provide evaluation class labels and achieve higher inter-annotator agreement. For the couples therapy dataset we will use a supervised evaluation procedure, even though the behavioral manifold has been trained in an unsupervised manner.

After we obtain the latent manifold representation for each frame, we use the Euclidean distance to find the closest ``reference frame'', which is the nearest frame among all the labeled frames from different couples. The leave-one-couple-out test procedure can ensure a fair evaluation where the speaker characteristics will not have any impact during testing. Further, we use the corresponding session behavior label as the reference frame's label. Then, we employ majority voting to generate session level labels from multiple frame level labels.

\subsection{Evaluation Method for OOD data}
Unlike the couples therapy data, our OOD do not have any labels. The evaluation data was selected however to reflect different behavioral styles. For instance as seen from Table \ref{tab:EoData}, a politician speaking during a debate is expected to be very different in behavioral style from a stand-up comedian, but similar to another politician. In this case we present the results of the clustering: what frame was close to what as a percentage. This percentage score implies the similarity between two audio frames.

\section{Results and Discussion}
\label{sec:discussion}
\subsection{Testing on within domain corpus}

\Pg{Baseline of couples' behavior classification} As a baseline system we use a nearest neighbor behavior classification in the acoustic feature space at the frame level, and similarly to  \ref{sec:IDeva} use majority voting to generate session level labels.
The results of this baseline classification method are shown in both Table \ref{tab:result_table_accept} and Table \ref{tab:multi-behaviors}, which are only slightly better than random guess. This result suggests that original acoustic features are not an effective candidate for behavior representation. Further training is needed in order to extract behavior information from high dimensional acoustic features.

\Pg{Comparison of within and OOD training}
In order to compare within and OOD training, we conduct experiment (1) and (2) on behavior code \emph{Acceptance}. To be consistent with our precious work\cite{xia2015dynamic, li2016_sparsely-connec}, a 20s frame size is chosen. Because of limited in-domain dataset (Ti) size, we build a neural network with only 2 hidden layers in that case. When training with  out-of-domain  data (To), since more training data is available, we employ a neural network with 5 hidden layers of  300, 200, 64, 200, 300 nodes respectively.
%Also, in Table \ref{tab:result_table_accept}, we also list supervised classification results by SD-DNN in \cite{li2016_sparsely-connec}.???? or Wei's results svm-hmm?
%% supervised in sd-dnn 75.36%

From the results in Table \ref{tab:result_table_accept}, both \emph{Ti} and \emph{To} training methods beat the performance of baseline, which shows that our audio2behavior framework is an effective way to project the signal on a more meaningful behavioral manifold in an unsupervised manner and reduce the feature dimensionality. As expected,  in-domain training performs better than that from OOD one. This is reasonable, since in terms of speech patterns and acoustic characteristics, there is a big gap between the movie and Couples therapy corpus, and importantly the couples data are far-field, low quality recordings while the movie data are usually higher quality signals. The mismatch of the training and test sets is minimal when both are from the same domain.

\Pg{Comparison of frame length} The OOD training result in Table \ref{tab:result_table_accept} is promising, especially for out-of-domain dataset, since we do not perform any pre-processing procedures, such as VAD or diarization. Because of that, as we mentioned in \ref{sec:pre-processing}, there should be some non-speech parts, such as background music, silence, as well as multiple sources of noise besides human speech. In addition, within a larger frame window size, it is also highly probable that speech regions within each window come from multiple speakers. Different speaker's characteristics in one frame window may contaminate behavior related acoustic representation. We try to find a proper way to improve the performance by reducing speaker characteristics. One approach is to reduce the length of frame window. We thus hypothesize that using a smaller window size the chance of single-speaker regions in each frame becomes higher and  thus it should improve the audio2behavior model performance by lowering acoustic complexity. This clearly assumes that the window, while smaller, is still long enough to capture the behavioral characteristics.

We employ experiment (2) on multiple behaviors with different frame lengths to verify this hypotheses. From the results in table \ref{tab:multi-behaviors}, we can see there is significant improvement, a 5\% absolute increase from 64.11\% to 69.11\% in terms of classification accuracy. Moreover, for all four behaviors, a consistent improvement is noted on 5s frame length acoustic feature. This shows that consistency within each acoustic speech frame region might be one critical issue in audio2behavior system, and encourages diarization as a front end pre-processing step. 
We should note here  that for complex human behavior annotation process, even for human annotators, the interannotator agreement can only reach about Krippendorff’s $\alpha=0.8$ \cite{tseng2016_couples-behavio}, and so the 69.11\% for a completely unsupervised method with just majority vote at the output is very encouraging. 

In general, these results are promising for communicative behavior quantification since we only utilize unlabeled, any-domain data and train in an unsupervised manner.

\begin{table}[t]
\centering
\caption{Classification accuracy (\%) of behavior acceptance}
\label{tab:result_table_accept}
\begin{tabular}{ccc}
\hline
Baseline & \begin{tabular}[c]{@{}c@{}}Train on Couples'\\ 20 s window size\end{tabular} & \begin{tabular}[c]{@{}c@{}}Train on Movies\\ 20 s window size\end{tabular} \\ \hline
57.5  & 69.29                                                                     & 66.43                                                           \\ \hline
\end{tabular}
\end{table}

\begin{table}[t]
\centering
\caption{Classification accuracy (\%) for behaviors with different frame window size}
\label{tab:multi-behaviors}
\begin{tabular}{cccc}
\hline
Behavior                                                   & Baseline & \begin{tabular}[c]{@{}c@{}}Train on \\ Movies \textbf{20 }s \\ window size\end{tabular} & \begin{tabular}[c]{@{}c@{}}Train on \\ Movies \textbf{5 }s \\ window size\end{tabular} \\ \hline
Acceptance                                                 & 57.50    & 66.43                                                                          & 68.57                                                                         \\ 
Blame                                                      & 55.00    & 61.07                                                                          & 71.78                                                                         \\ 
Negtivity                                                  & 63.93    & 63.93                                                                          & 69.64                                                                         \\ 
Positivity                                                 & 51.07    & 65.00                                                                          & 66.43                                                                         \\
Average & 56.88    & 64.11                                                                          & \textbf{69.11}                                                                         \\ \hline
\end{tabular}
\end{table}

\subsection{Testing on OOD corpus}
\begin{figure}[t]
  \centering
  \includegraphics[width=\linewidth]{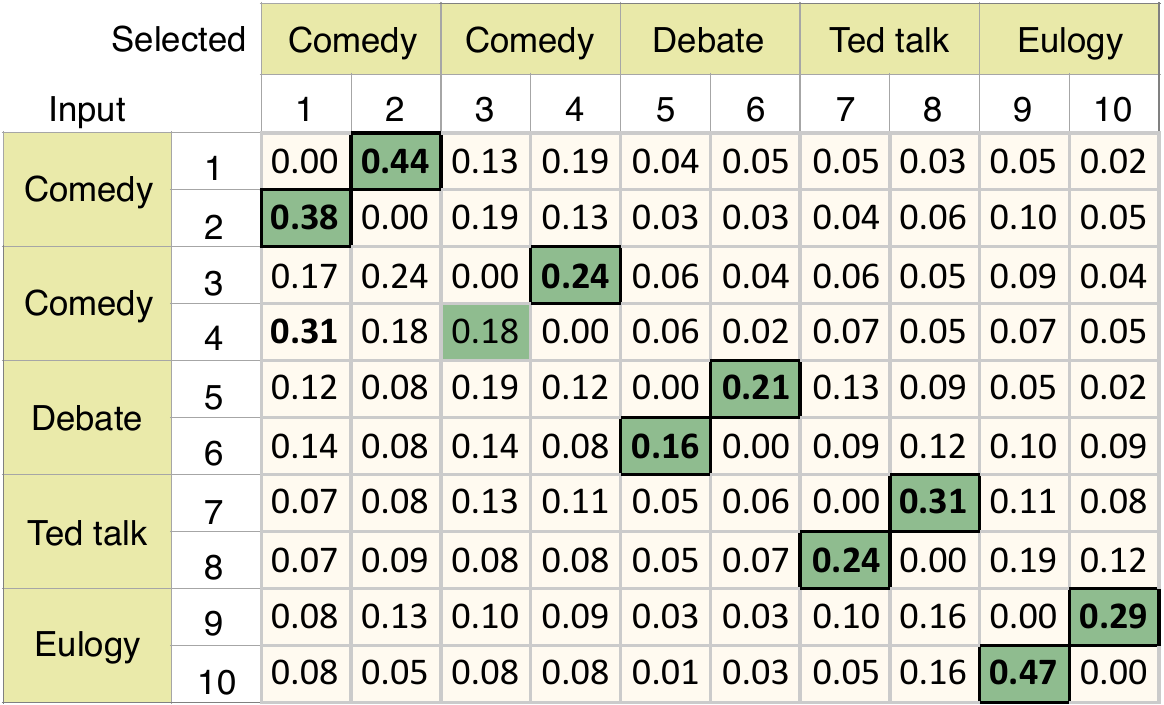}
  \caption{{Behavior scenario similarity evaluation results}}
  \label{fig:table}
\end{figure}

As mentioned in section \ref{sec:data}, we collect OOD test dataset from different scenarios listed in Table \ref{tab:EoData}. In each scenario, two audio files are collected from different speakers. We use normalized percentage score to evaluate behavior similarity. The score is calculated by dividing number of nearest frames in each selected scenario by the number of total frames of input audio. Results are shown in Figure \ref{fig:table}.

Ideally, audio from similar scenarios should exhibit high similarity with each other, and a lower score should be assigned between less related scenarios. We find that 9 out of 10 audio samples are classified as we hoped based on majority vote on frame level clustering. Moreover, besides classification, results show behavior similarity with details: audio2behavior can show behavior similarity under different degrees. For example, we can see that 44\% of data from George Carlin are identified as similar to Richard Pryor, where both comedians employ an angry tone in their stand-up comedy and 19\% and 13\% come from Steve Hofstetter and Jim Gaffigan, also comedians that employ a milder tone in their routines. Less than 5\% of the data are associated with any of the other conditions. All these results show promising behavioral quantification of our audio2behavior model.

\section{Conclusion and future work}
\label{sec:Conclusion}

Data sparsity is always a critical issue in behavior related studies. Behavior recognition research suffers from expensive data annotation process and low inter-annotator agreement, which also limits the performance of automated behavior recognition system. Compared with previous existing supervised behavioral recognition in BSP domain, our audio2behavior provides another possible solution candidate: transfer out of domain knowledge into training, then adapt the model into domain applications. 
This unsupervised training approach of vectorizing abstract behavior from audio and then obtaining better behavioral quantification in manifold shows auspicious results and applications in behavioral signal processing domain.

In the future, inspired by results of this paper, we plan to employ VAD and diarization into the front end to better improve the training of the audio2behavior model. This will reduce speaker characteristics and acoustic complexity in behavior representation by allowing us to do speaker-specific normalizations.
Alternatively we can employ the speaker-distinct regions but in a joint and unsupervised manner learn both a speaker and behavioral manifold.

Moreover, unsupervised behavior representation models can be also employed into a range of applications for which training data are unavailable, by quickly allowing out-of-domain bootstrapping.

%\vfill
%\pagebreak

%\vfill\pagebreak

%\ninept

\bibliographystyle{IEEEbib}
\bibliography{refs}

\end{document}